\documentclass[runningheads]{llncs}
\usepackage{amsmath}
\usepackage{algorithm}
\usepackage{booktabs} 
\usepackage{multirow}
\usepackage{hyperref}
\usepackage{graphicx}
\usepackage{subcaption} 
\usepackage{hyperref}

\usepackage{amsmath}
\usepackage{amssymb}
\usepackage{algorithm}
\usepackage{algpseudocode}
\usepackage{longtable}
\usepackage[english]{babel}
\usepackage[T1]{fontenc}
\usepackage{graphicx}
\begin{document}
\raggedbottom
\title{Q-SafeML: Safety Assessment of Quantum Machine Learning via Quantum Distance Metrics}
%
%
\author{Oliver Dunn\inst{1}\orcidID{0000-1111-2222-3333} \and
Koorosh Aslansefat\inst{1}\orcidID{1111-2222-3333-4444} \and Yiannis Papadopoulos\inst{1}\orcidID{0000-0001-7007-5153}}
\authorrunning{O. Dunn et al.}
%
\institute{University of Hull, Hull East Yorkshire HU6 7RX, United Kingdom 
\email{o.dunn2-2021@hull.ac.uk}\\
\url{https://plennock.github.io/Honours-Stage-Project/}}
\titlerunning{QSafeML: Safety Evaluation of Quantum ML Models}
\maketitle              
\begin{abstract}
The rise of machine learning in safety-critical systems has paralleled advancements in quantum computing, leading to the emerging field of Quantum Machine Learning (QML). While safety monitoring has progressed in classical ML, existing methods are not directly applicable to QML due to fundamental differences in quantum computation. Given the novelty of QML, dedicated safety mechanisms remain underdeveloped. This paper introduces Q-SafeML, a safety monitoring approach for QML. The method builds on SafeML, a recent method that utilizes statistical distance measures to assess model accuracy and provide confidence in the reasoning of an algorithm. An adapted version of Q-SafeML incorporates quantum-centric distance measures, aligning with the probabilistic nature of QML outputs. This shift to a model-dependent, post-classification evaluation represents a key departure from classical SafeML, which is dataset-driven and classifier-agnostic. The distinction is motivated by the unique representational constraints of quantum systems, requiring distance metrics defined over quantum state spaces. Q-SafeML detects distances between operational and training data addressing the concept drifts in the context of QML. Experiments on QCNN and VQC Models show that this enables informed human oversight, enhancing system transparency and safety.

\end{abstract}
\section{Introduction}

Safety monitoring in machine learning (ML) is crucial for safety-critical domains such as healthcare \cite{adepoju2019}, transportation \cite{ahmed2025}, and energy \cite{benedetti2019}. One such method, \textit{SafeML}, uses parametric distance measures to assess how closely a model’s predictions align with real-world data, helping detect weaknesses and uncertainty before or during deployment \cite{paleyes2022mlchallenges}.
\\
\\
Originally developed to address out-of-distribution detection (OODD) in ML classifiers \cite{liang2020odin,kim2024ood}, SafeML now intersects with a growing field: quantum machine learning (QML). QML leverages quantum computing principles such as superposition and entanglement to enhance ML tasks like classification and regression \cite{schuld2020,gilfuster2018}.
\\
\\
As classical models transition to quantum environments, safety tools must also adapt. Yet, safety remains underexplored in QML, despite the well-known fragility of quantum systems and their sensitivity to noise \cite{xu2021mlsafety}. This work introduces \textit{Quantum SafeML}, an adaptation of SafeML that integrates quantum-specific statistical distance measures for safety monitoring in quantum settings \cite{ibmroadmap2024,google2024}.
\\
\\
Here, “safety” refers to identifying unreliable or error-prone behavior in ML models—particularly incorrect or uncertain predictions. The goal is not formal correctness, but improved transparency and early risk detection during deployment.

\subsection{Aims and Objectives}

The goal is to develop a QML-compatible version of SafeML using quantum distance metrics to monitor prediction accuracy and model reliability. Key objectives:
\begin{itemize}
    \item Identify suitable quantum distance measures.
    \item Evaluate their fit within the SafeML framework.
    \item Implement them on quantum datasets.
    \item Use them to assess QML model performance.
    \item Evaluate the overall effectiveness of Quantum SafeML.
\end{itemize}

\subsection{Research Questions}
\begin{itemize}
    \item RQ1: How effective is SafeML at detecting labeling errors in QML?
    \item RQ2: Can it adapt across various QML implementations?
    \item RQ3: Which QML applications benefit most from such monitoring?
\end{itemize}

\subsection{Scope and Limitations}

QML is expected to impact future technologies, especially in safety-critical areas \cite{eu2024aiact}. However, safety monitoring tools for QML remain immature. This work adapts SafeML to quantum contexts to provide a foundation for more robust safety practices and contributes to the validation of emerging quantum technologies \cite{paleyes2022mlchallenges,xu2021mlsafety}.

Due to limited access to quantum hardware, experiments used simulators such as Qiskit, PennyLane, and Cirq. This project used Qiskit for its built-in QML tools and extensive IBM documentation. While simulators don't fully replicate quantum noise, they are sufficient for proof-of-concept validation. The focus is on performance monitoring in classification tasks, not on outperforming classical models.

Unlike classical SafeML, this version is model-dependent and monitors classifier behavior post-decision, rather than input distribution drift. This shift accommodates quantum-specific outputs, such as density matrices, which are incompatible with classical SafeML’s assumptions.

\subsection{Paper Structure}

Section 2 reviews related work in QML and SafeML. Section 3 outlines the research gap. Section 4 presents Quantum SafeML, followed by experiments in Section 5. Results and discussion appear in Section 6, with conclusions in Section 7.

\section{Related Work}

\subsection{Quantum Computing and Machine Learning}

Quantum computing introduces a new computational paradigm based on quantum mechanical principles like superposition and entanglement \cite{ladd2010}. Although still in early development, recent research explores how quantum algorithms can support or enhance machine learning tasks \cite{benedetti2019,adepoju2019,ahmed2025}. Prominent approaches include quantum kernels, Boltzmann machines, and variational quantum circuits, which may offer expressivity or training benefits in specific contexts \cite{schuld2020,gilfuster2018}.

Quantum information theory contributes new tools to ML, such as density matrices and entropy measures, offering alternative ways to model uncertainty and correlations \cite{ahmed2025}. While widespread QML deployment is limited by hardware, its theoretical influence is shaping developments in model architecture and learning theory \cite{ibmroadmap2024,google2024}.

\subsection{Out-of-Distribution Detection}

Out-of-distribution detection (OODD) addresses the risk that models encounter test data different from their training distribution \cite{paleyes2022mlchallenges}. This issue is especially critical in domains like healthcare, autonomous driving, and security, where undetected distributional shifts can lead to serious failures \cite{xu2021mlsafety,eu2024aiact}.

Recent techniques include Bayesian uncertainty quantification, deep generative models for density estimation, and contrastive learning \cite{liang2020odin,kim2024ood}. Many methods build on geometric properties of feature spaces or assume smoothness and separability in the learned representations \cite{kim2024ood}. There's also increasing focus on robustness guarantees and formal generalization metrics under shift, drawing on statistical learning and information theory \cite{zolfagharian2025smarla}.

Challenges remain, such as benchmark definition, calibration under shift, and understanding the links between OOD behavior and training-time regularization \cite{paleyes2022mlchallenges,xu2021mlsafety}. Ongoing research continues to explore the balance between novelty sensitivity and robustness to natural variation \cite{liang2020odin}.

\subsection{Runtime Safety Monitoring and SafeML}

Runtime monitoring and human oversight are essential for safe ML deployment. SafeML supports this by applying statistical methods to assess model reliability during operation. It evaluates how much new input data deviates from the training distribution using empirical cumulative distribution function (ECDF)-based distance metrics. This helps estimate potential accuracy drops and prompts human intervention when data shifts are detected \cite{aslansefat2020safeml}.

SafeML is one of few techniques cited in standards like Germany's DIN SPEC 92005 for ML uncertainty quantification \cite{DIN92005}. It has been extended for image classification via bootstrapping methods \cite{aslansefat2021toward}, and to time-series and regression tasks using new robustness metrics \cite{akram2022stadre}. Threshold tuning remains a challenge; automated threshold adjustment methods have been proposed \cite{farhad2022keep}.

Applications span fields such as intrusion detection \cite{aslansefat2020safeml}, autonomous driving \cite{bergler2022case,aslansefat2021toward}, robotics \cite{cho2022online,aslansefat2022safedrones}, and UAV inspection \cite{kabir2022combining}. Recent work extends SafeML to analyze internal neural network layers for more precise distribution shift detection \cite{farhad2023scope}. The SMILE extension builds on SafeML to offer model-agnostic explanations through empirical distance measures \cite{aslansefat2023explaining}.

\section{Research Gap and Problem Definition}

While classical safety-monitoring techniques like SafeML have shown effectiveness in identifying classifier confusion through statistical distance metrics, their direct application to quantum systems remains unexplored. Key research gaps include:
\begin{itemize}
    \item Absence of a quantum-adapted SafeML framework.
    \item Limited error estimation protocols for QML.
    \item No empirical method for analyzing false positives/negatives in QML using distribution-based techniques.
    \item Existing quantum distance metrics have not been integrated into safety monitoring.
\end{itemize}

Classical SafeML operates on deterministic predictions and employ metrics (e.g., Kolmogorov-Smirnov, Wasserstein) which are incompatible with quantum representations like density matrices. The difficulty is that quantum models de facto produce probabilistic outputs, requiring different representations and tools. As such, classical SafeML fails to account for:
\begin{enumerate}
    \item Quantum data superposition and representation via density matrices.
    \item Probabilistic outputs from quantum circuits.
    \item Incompatibility of classical statistical distances with quantum data formats.
\end{enumerate}

To address the above, we explore how SafeML can be meaningfully adapted to QML architectures using quantum-compatible statistical distance metrics. A core requirement is transforming predictions into density matrices:
\begin{equation}
\rho = \sum_i p_i | \psi_i \rangle \langle \psi_i |
\end{equation}
where \( \rho \) encodes a quantum state as a probabilistic mixture of pure states \( | \psi_i \rangle \). We propose, and evaluate a solution for meaningful safety monitoring of Quantum ML. Our study is limited to classification problems. Although such an approach may potentially generalize to other learning paradigms, regression and reinforcement learning are beyond the current scope.

\section{The Quantum SafeML Method}

The Quantum SafeML method extends the classical SafeML framework to accommodate the probabilistic nature of quantum computations. This is achieved by integrating quantum-specific distance metrics, namely trace distance, fidelity, Bures distance, and quantum relative entropy, to assess the similarity between quantum states and evaluate the reliability of QML models.

\subsection{Trace Distance}

Trace distance quantifies the distinguishability between two quantum states, providing a measure of how well one can differentiate between them. For density matrices \( \rho \) and \( \sigma \), it is defined as:

\begin{equation}
D(\rho, \sigma) = \frac{1}{2} \operatorname{Tr} \left| \rho - \sigma \right|
\end{equation}

Here, \( \left| \rho - \sigma \right| \) denotes the trace norm of the matrix difference. A trace distance of 0 indicates identical states, while a value of 1 signifies completely distinguishable states~\cite{nielsen2002quantum}.

\subsection{Fidelity}

Fidelity measures the similarity between two quantum states, reflecting the probability that one state will pass a test to identify as the other. For density matrices \( \rho \) and \( \sigma \), fidelity is given by:

\begin{equation}
    F(\rho, \sigma) = \left( \operatorname{Tr} \sqrt{\sqrt{\rho} \, \sigma \, \sqrt{\rho}} \right)^2
\end{equation}

A fidelity of 1 implies identical states, whereas 0 indicates orthogonal, i.e.  completely distinct, states~\cite{jozsa1994fidelity}.

\subsection{Bures Distance}

Derived from fidelity, the Bures distance provides a metric for the geometric distance between quantum states:

\begin{equation}
D_B(\rho, \sigma) = \sqrt{2 \left( 1 - \sqrt{F(\rho, \sigma)} \right)}
\end{equation}

This metric is particularly useful in quantum information theory for comparing mixed quantum states~\cite{bures1969extension,uhlmann1976transition}.

\subsection{Quantum Relative Entropy}

Quantum relative entropy measures the distinguishability between two quantum states, analogous to the classical Kullback-Leibler divergence. For density matrices \( \rho \) and \( \sigma \), it is given by:

\begin{equation}
S(\rho \| \sigma) = \operatorname{Tr} \left( \rho \log \rho - \rho \log \sigma \right)
\end{equation}

This quantity represents the information loss when approximating \( \rho \) with \( \sigma \)~\cite{vedral2002role}.

Quantum SafeML employs these distance metrics to compare two sets: one containing  misclassified predictions for a specific  label, and the other containing  correctly classified ones. The computed  distance metric can then be evaluated against  the model’s true accuracy, potentially within an optimization framework.

This process involves training models on a dataset as usual. During validation, classification outcomes are recorded and stored in a suitable  format (e.g., density matrix or NumPy array). These sets are then analysed  using the selected  metrics and compared with the model's accuracy to yield a final assessment of classifier safety.

\subsection{Differentiation from Classical SafeML}
While Q-SafeML draws conceptual inspiration from the classical SafeML framework, it represents a significant methodological departure. Classical SafeML is fundamentally model-agnostic: it operates on data distributions alone, comparing statistical differences between the training set and new inputs to detect distributional shifts without direct reference to the classifier’s internal behavior.

In contrast, Q-SafeML is inherently model-dependent. Rather than evaluating input drift, it assesses the reliability of a quantum classifier’s outputs by comparing sets of correctly and incorrectly classified predictions using quantum distance metrics. This shift reflects the probabilistic and state-based nature of quantum machine learning, where outputs are quantum states (or statistical mixtures thereof), and the notion of “distance” must account for this complexity.

This adaptation is necessary because quantum systems encode both data and results in formats that are incompatible with classical statistical techniques. Therefore, while Q-SafeML preserves SafeML’s core philosophy—detecting unsafe or unreliable model behavior using statistical comparisons—it operates at a fundamentally different layer of the machine learning pipeline: post-classification rather than pre-deployment.

\begin{figure}[H]
    \centering
    \includegraphics[width=0.9\linewidth]{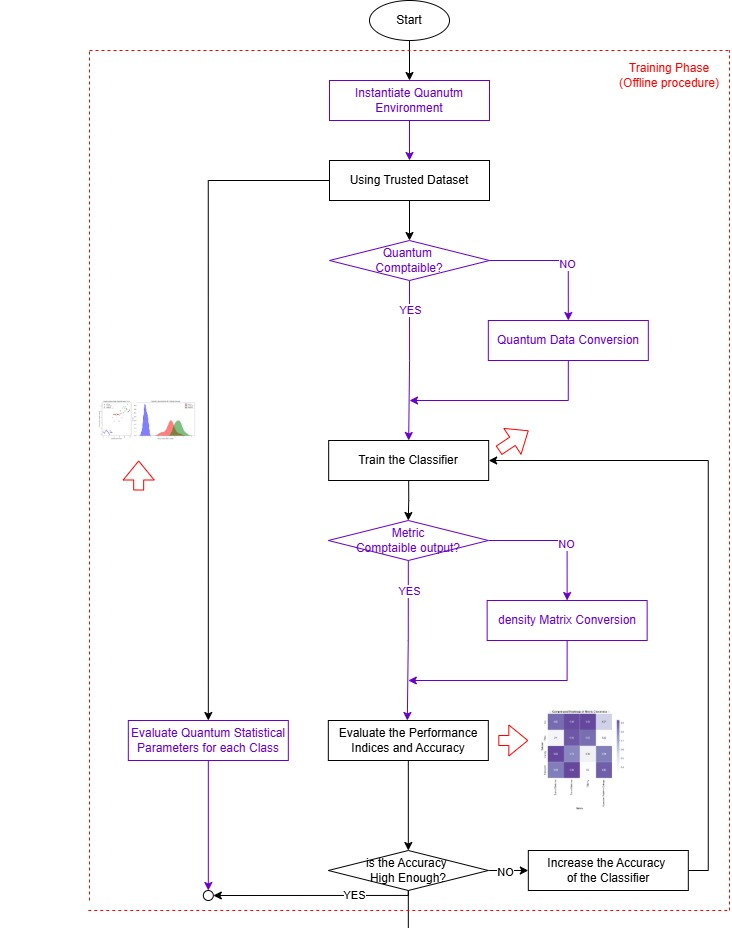}
    \caption{Proposed Quantum SafeML method (training phase)}
    \label{flowchart}
\end{figure}

\begin{figure}[H]
    \centering
    \includegraphics[width=1\linewidth]{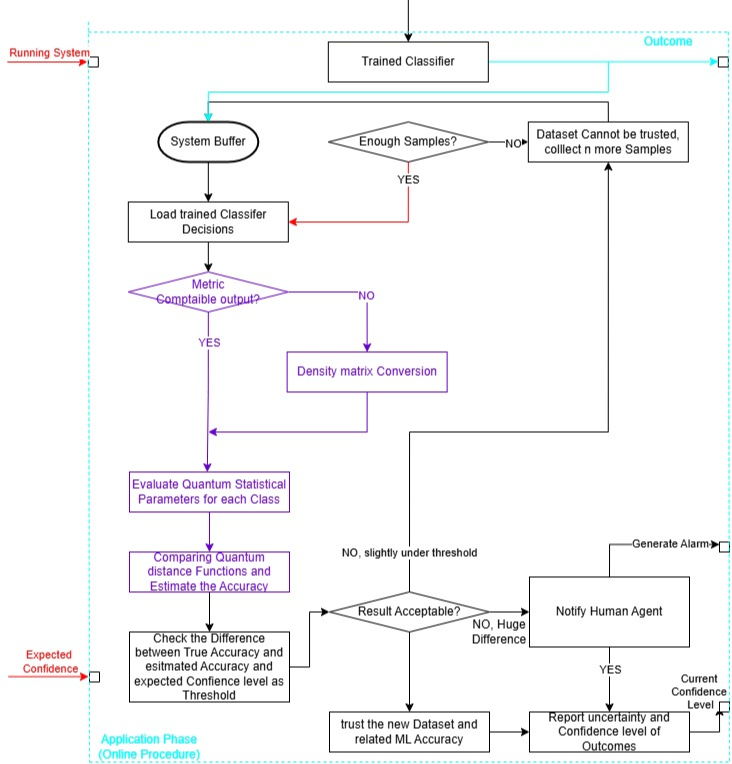}
    \caption{Proposed Quantum SafeML method (online phase)}
    \label{fig:enter-label}
\end{figure}

\section{Case Studies}

\subsection{Quantum Environment Instantiation with Qiskit}

Qiskit was chosen for quantum environment instantiation due to its robust simulation tools, active support community, and access to IBM quantum hardware. While alternatives like PennyLane focus on quantum machine learning, Qiskit offers a dedicated QML library, broader algorithmic support, and hybrid model compatibility. Recent updates, though introducing a steeper learning curve, have made Qiskit more future-proof and reliable for experimentation.
\\
\\
The simulated environment was built using Qiskit documentation and example notebooks, enabling accurate modeling of quantum behavior such as superposition and entanglement. Core components like quantum gates, circuit construction, and backend configuration were included to support scalable QML testing. This setup offers a controlled yet realistic platform for developing and evaluating the Quantum SafeML framework.
\\
\\
To assess the effectiveness of Quantum SafeML, we applied it to two quantum models: a Variational Quantum Classifier (VQC) evaluated on toy datasets and a Quantum Convolutional Neural Network (QCNN) applied to handwritten digit classification.

\subsection{Quantum SafeML on a VQC with Toy Datasets}

The VQC model, originally built for the Iris dataset, was extended to classify the Wine dataset (with 13 numerical features) as well as several synthetic datasets with randomised features. Quantum SafeML was then used to evaluate model performance using Bures distance, Trace distance, Fidelity, and Quantum relative entropy


\begin{table}[h]
    \centering
    \renewcommand{\arraystretch}{1.2}
    \resizebox{\textwidth}{!}{
    \begin{tabular}{lccccc}
        \toprule
        \textbf{Dataset} & \textbf{Bures Distance} & \textbf{Trace Distance} & \textbf{Fidelity} & \textbf{Quantum Relative Entropy} & \textbf{True Accuracy} \\
        \midrule
        Iris      & 0.7482 & 0.5000 & 0.4677 & 1.2396 & 0.5333 \\
        Wine      & 0.9036 & 0.3701 & 0.2517 & 1.0000 & 0.3056 \\
        Family    & 0.3352 & 0.0535 & 0.8893 & 0.2424 & 0.4250 \\
        Transport & 0.3627 & 0.1299 & 0.8714 & 0.2823 & 0.1406 \\
        \bottomrule
    \end{tabular}
    }
    \caption{Quantum Metric Comparisons Across Datasets the VQC trained on.}
    \label{tab:quantum_metrics}
\end{table}

To assess how well quantum distance metrics reflect model reliability, we computed Pearson correlation coefficients between each metric and true classification accuracy. Quantum relative entropy showed the highest (moderate) correlation with accuracy (r = 0.54), followed by trace distance (r = 0.48). Fidelity and Bures distance showed weaker correlations. However, due to the limited number of datasets, these results are not statistically significant and should be interpreted cautiously.

\begin{figure}
    \centering
    \begin{subfigure}{0.49\linewidth}
        \centering
        \includegraphics[width=\linewidth]{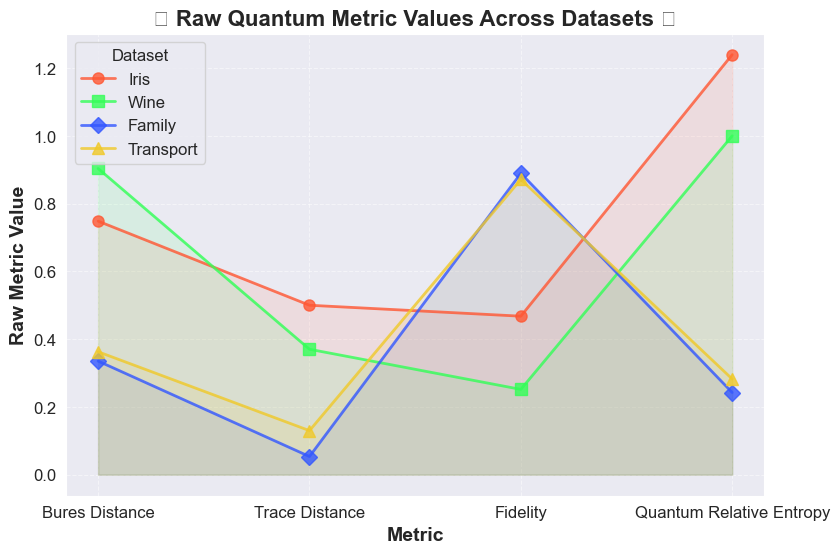}
        \caption{Raw Quantum Metric Values Across Datasets}
        \label{fig:metric-values}
    \end{subfigure}
    \hfill
    \begin{subfigure}{0.49\linewidth}
        \centering
        \includegraphics[width=\linewidth]{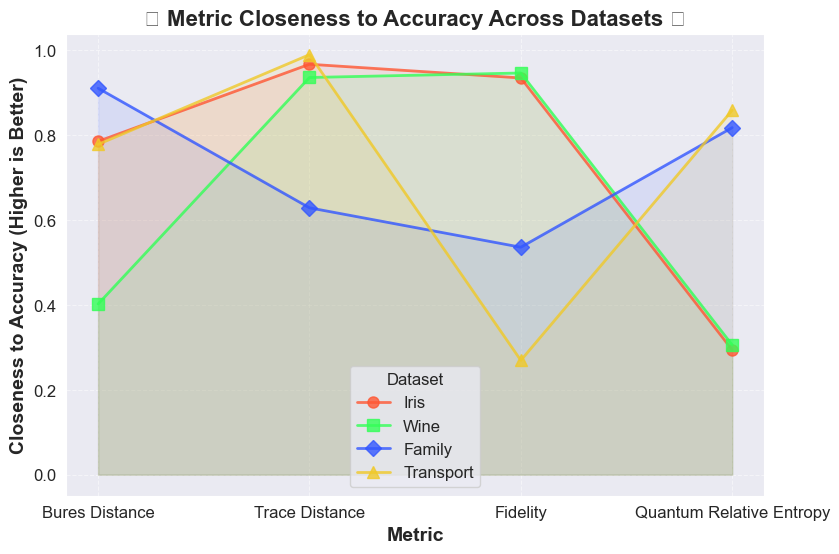}
        \caption{Metric Closeness to Accuracy Across Datasets}
        \label{fig:metric-accuracy}
    \end{subfigure}
    \caption{Comparison of Quantum Metric Values Across each of the Datasets the VQC trained on, and the overall model accuracy for each of these.}
    \label{fig:combined-metrics}
\end{figure}

\begin{figure}[H]
    \centering
    \includegraphics[width=0.5\linewidth]{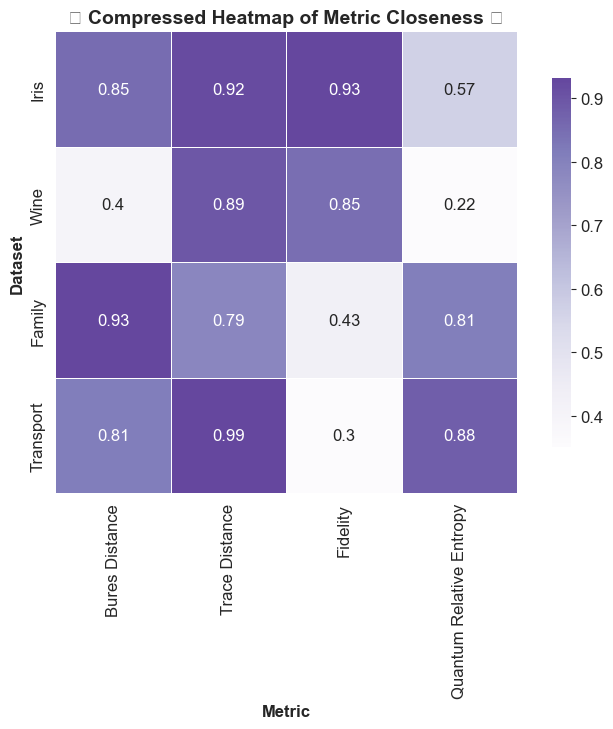}
    \caption{Heatmap of Metric Values Across Different Datasets the VQC trained on.}
    \label{fig:VQC-heatmap}
\end{figure}


Key findings from the analysis indicate:  

\begin{itemize}
    \item A quantum relative entropy value greater than 1 may indicate potential issues with matrix shape or structural integrity.

    \item The Wine dataset exhibits a notably lower accuracy of 30.6\% and simultaneously higher metric distances, signalling a notable discrepancy in classification reliability.

    \item This discrepancy is visualised in Figure~\ref{fig:combined-metrics}, which shows the comparative metric values across datasets. Figure~\ref{fig:VQC-heatmap} further illustrates how these metric variations correlate with the classifier’s predictive performance, reinforcing the observed performance degradation on the Wine dataset.
\end{itemize}

\subsection{Quantum SafeML on a QCNN with Digit Dataset}

The QCNN was applied to classify 8×8 handwritten digit images with dimensionality reduced via PCA to 6 principal features. Key findings include:

\begin{figure}
    \centering
    \begin{subfigure}{0.49\linewidth}
        \centering
        \includegraphics[width=\linewidth]{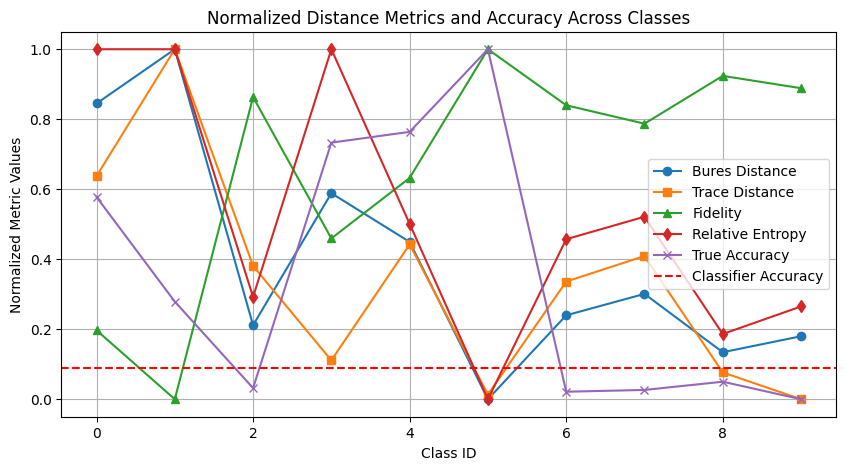}
        \caption{Metric Values for QNN Classes}
        \label{fig:metric-values-QNN}
    \end{subfigure}
    \hfill
    \begin{subfigure}{0.49\linewidth}
        \centering
        \includegraphics[width=\linewidth]{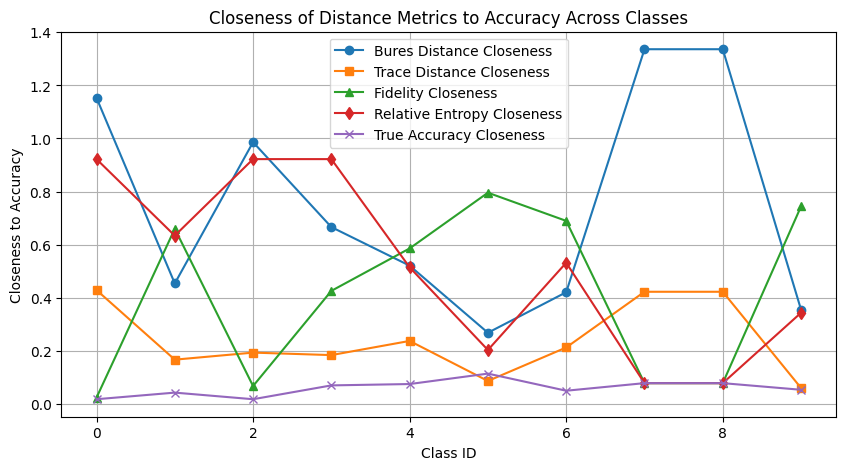}
        \caption{Metric Accuracy Closeness in QNN}
        \label{fig:metric-accuracy-closeness-QNN}
    \end{subfigure}
    \caption{Comparison of distance Metric Values through each of the different classes in the digits dataset, and the closeness to model accuracy for each of these.}
    \label{fig:QNN-metric-comparison}
\end{figure}

\begin{figure}
    \centering
    \includegraphics[width=0.8\linewidth]{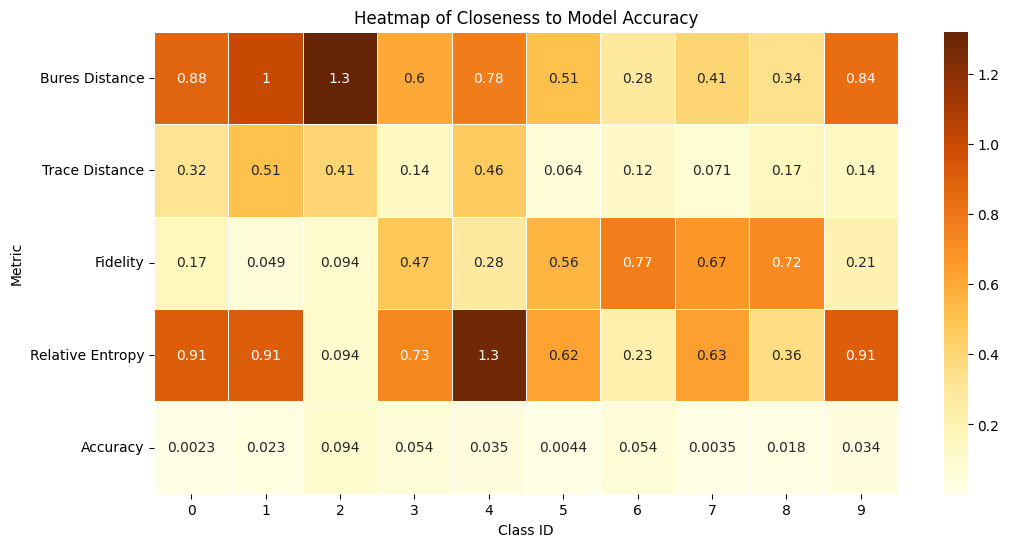}
    \caption{Heatmap showing how each metrics closeness to model accuracy varies across the subsets derived from each class label.}
    \label{fig:QNN heatmap}
\end{figure}


\begin{itemize}
    \item A high variance in the Bures distance was observed for classes 7 and 8, as illustrated in Figure~\ref{fig:metric-values-QNN}. This suggests instability or heightened uncertainty in the model’s internal representation of these classes. 
    \item The heatmap presented in Figure~\ref{fig:QNN heatmap} reveals distinct class-specific patterns in quantum metric values, offering further insights into the model’s interpretability and behaviour across the digit classes.
\end{itemize}

\subsection{Exploring Threshold Values with Quantum Kernel Methods}

\begin{figure}[h]
    \centering
    \includegraphics[width=1\linewidth]{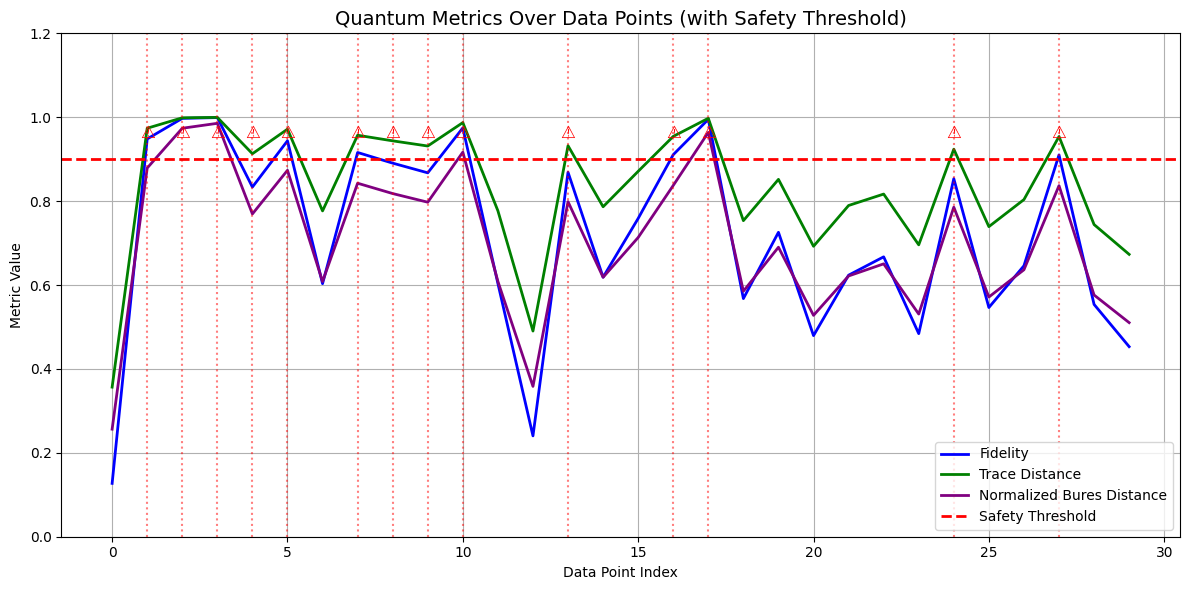}
    \caption{Kernel-based metric values compared to threshold; exceeding the line suggests lower model confidence.}
    \label{thresholding}
\end{figure}


The thresholding framework operates by first normalizing the relevant safety metrics: the Bures distance is divided by \(\sqrt{2}\) and the complement of fidelity (i.e., \(1 - \text{fidelity}\)) is computed. After normalisation, the system flags any samples whose metric values exceed predefined thresholds, marking them as potentially unsafe. This behaviour is illustrated in Figure~\ref{thresholding}, which shows instances where the threshold has been breached.

\section{Discussion}
Quantum SafeML has demonstrated its ability to  identify weaknesses in QML, particularly across both VQC and QCNN architectures, through application of quantum distance  metrics. This section discusses the effectiveness, challenges, and implications of the approach.

\subsection{Effectiveness of SafeML in Quantum Machine Learning}

Quantum SafeML proved effective in delivering detailed performance assessments across multiple distance metrics, allowing for meaningful comparisons with classifier accuracy. 

For VQC models, SafeML performed strongly   across diverse datasets, despite tolerating some inconsistencies, e.g., from quantum relative entropy in sklearn toy datasets, where misclassified data could inflate results. Despite some shortcomings, the metrics provided valuable insight into model weaknesses nd uncertainty.
\\
\\
In the case of QCNNs, SafeML the method encountered greater complexity. The architectural intricacies of  QCNNs made SafeML application more challenging, requiring careful configuration. However, it still remained capable of identifying prediction errors, offering insights into model behavior.
\\
\\
These findings suggest that among the evaluated metrics, quantum relative entropy and trace distance show the highest (moderate) correlation with model accuracy, indicating some potential as reliability indicators. In contrast, fidelity and Bures distance demonstrated weaker or inconsistent relationships. Given the small sample size and lack of statistical significance, these insights should be interpreted cautiously, and future work should explore combining metrics or integrating confidence thresholds for more robust monitoring.

\subsection{Metric Selection and Application}

Choosing the right metric for Quantum SafeML is heavily dependent on the specific goals of the safety validation process. The following insights were derived from the experiments:

\begin{itemize}
    \item \textbf{Trace Distance} was the most consistent and interpretable metric, reliably capturing classifier confusion across various VQC tasks.
    \item \textbf{Bures Distance} performed well in noisy conditions and mixed quantum states, i.e. conditions expected in real quantum hardware, being more sensitive to noise.
    \item \textbf{Fidelity} excelled in generative tasks or for measuring state overlap. However, it was sensitive to encoding quirks and entanglement issues.
    \item \textbf{Quantum Relative Entropy} proved useful for asymmetric error evaluation and privacy-related divergence, though it required  careful handling to avoid invalid or misleading results.
\end{itemize}
When used together, the metrics offered complementary insights. For instance, cases where fidelity and trace distance disagreed—such as in the Wine dataset—often corresponded to ambiguous classification boundaries or poorly separated feature spaces. This divergence can serve as a red flag, suggesting internal inconsistencies in the model's confidence. In contrast, convergence of all metrics typically aligned with high-accuracy predictions and more stable output states. This highlights the value of a multi-metric approach over relying on any single distance measure in isolation.
\\
\\
No single metric emerged as universally optimal. Instead, using multiple metrics in parallel, analyzing their convergence or divergence, provided a more holistic understanding of classifier performance.

\subsection{Quantum Hardware Considerations}
The study primarily relied on simulators, which assume idealized conditions unless explicitly configured otherwise. As such, they do not capture the full range of errors and noise present in current quantum hardware, such as gate infidelity, crosstalk, and decoherence. While Bures distance and quantum relative entropy are sensitive to such distortions and may behave differently under real hardware constraints, our simulation results likely present an optimistic view.
\\
\\
To mitigate this, future work will incorporate noise models available in Qiskit to simulate realistic hardware conditions more faithfully. Long-term, we aim to evaluate Q-SafeML on IBM Q and other public quantum backends. Additionally, reducing circuit depth and optimizing encodings may enhance the method’s robustness on Noisy Intermediate-Scale Quantum (NISQ) devices.

\subsection{Limitations and Challenges}

A major challenge in applying SafeML lies in handling multiclass classification problems. Encoding valid quantum states across multiple output classes adds complexity, especially when mapping model predictions to quantum systems.

Additionally, SafeML was applied post-hoc, i.e.  after model predictions were made. This may have limited its potential. A more integrated approach, potentially using continuous monitoring during training or metric-based loss functions, could offer more robust results.

Finally, practical issues inherent in live quantum environments, such as latency, calibration, and noise, were not fully accounted for in this study due to the reliance on simulators. These factors are likely to impact performance in real-world QML applications.

\subsection{Evaluation of Initial Research Questions}

\subsubsection*{RQ1: How effective is SafeML in detecting labelling errors in QML?}
Quantum SafeML proved effective in identifying labelling errors in QML classification tasks. By applying distance metrics to density matrix representations, it reliably highlighted mislabelled instances and quantified their occurrence. For example, the Wine dataset exhibited lower accuracy and high divergence in metric values (see Table~\ref{tab:quantum_metrics} and Figure~\ref{fig:combined-metrics}), indicating inconsistencies linked to potential labelling or model limitations.

\subsubsection*{RQ2: Is it adaptable to different QML implementations?}
While this study focused solely on classification models, Quantum SafeML was tested across various architectures — specifically the VQC and QCNN (see Sections~4.2 and~4.3). The flexibility of using multiple distance metrics enabled adaptation to datasets of varying dimensionality and structure. This adaptability is further supported by class-specific patterns observed in the QCNN analysis (Figure~\ref{fig:QNN heatmap}), demonstrating its potential for generalisation to other QML contexts.

\subsubsection*{RQ3: Which QML applications benefit most from such monitoring?}
Among the tested models, the QCNN used for image classification exhibited the clearest benefit from Quantum SafeML. As highlighted in Figures~\ref{fig:metric-values-QNN} and~\ref{fig:QNN heatmap}, distinct quantum metric behaviours across digit classes allowed targeted error identification. This aligns with findings in Section~4.3, and mirrors the strengths of classical SafeML in computer vision applications.

\section{Conclusion}

This study evaluated the performance of different quantum distance metrics within the Quantum SafeML framework. The findings indicated that each metric has specific strengths and weaknesses depending on the classifier type and dataset. Trace distance emerged as the most stable, while Bures distance was most suited to noisy settings. Fidelity was well-suited for generative tasks, and Quantum Relative Entropy was most effective for evaluating asymmetric errors.

As quantum computing continues to develop, the safety and reliability of quantum machine learning models will remain a critical concern. Methods like Quantum SafeML, combined with careful metric selection, could play a vital role in making quantum models more transparent, trustworthy, and robust for real-world deployment.

\section{Code Availability}
Regarding the research reproducibility, codes and functions supporting this paper are published online at GitHub: \\ \href{https://github.com/Plennock/Honours-Stage-Project}{https://github.com/Plennock/Honours-Stage-Project}.

\bibliographystyle{splncs04}
\bibliography{References.bib}

\end{document}